\documentclass[11pt]{article}
\usepackage{coling2020}

\usepackage{times}
\usepackage{url}
\usepackage{latexsym}
\usepackage{covington}
\usepackage{multirow}
\usepackage{rotating}
\usepackage[round, authoryear]{natbib}
\usepackage{xcolor}

\usepackage{graphicx}
\usepackage{caption, subcaption}
\graphicspath{ {./figures/} }
\usepackage{pgfplots}
\usepackage{wrapfig}

\usepackage{booktabs}

\usepackage{enumitem}% http://ctan.org/pkg/enumitem

\usepackage{amsmath}
\usepackage{amssymb}

\usepackage[T1,T2A]{fontenc}
\usepackage[utf8]{inputenc}
\usepackage[russian,english]{babel}

\usepackage[T1]{fontenc}

\usepackage{hyperref}

\colingfinalcopy % Uncomment this line for the final submission

\title{Rediscovering the Slavic Continuum in Representations Emerging \\ from Neural Models of Spoken Language Identification}

  \author{Badr M. Abdullah  \hspace{0.75cm} Jacek Kudera  \hspace{0.75cm} Tania Avgustinova \\
  \textbf{Bernd Möbius \hspace{0.75cm} Dietrich Klakow } \\ %\vspace{0.5}
         Department of Language Science and Technology (LST)  \\ Saarland Informatics Campus, Saarland University, 66123 Saarbrücken, Germany \\ %$\dagger$ Saarland Informatics Campus \textsuperscript{1,2} \\ Spoken Language Systems (LSV) \\ 
        Corresponding author: \normalsize{\texttt{babdullah@lsv.uni-saarland.de}} 
}

\date{}

\begin{document}
\maketitle
\begin{abstract}
Deep neural networks have been employed for various spoken language recognition tasks, including tasks that are multilingual by definition such as spoken language identification. In this paper, we present a neural model for Slavic language identification in speech signals and analyze its emergent representations to investigate whether they reflect objective measures of language relatedness and/or non-linguists' perception of language similarity. While our analysis shows that the language representation space indeed captures language relatedness to a great extent, we find perceptual confusability between languages in our study to be the best predictor of the language representation similarity.
\end{abstract}

% --------------------------------------------------------------
\section{Introduction}
% --------------------------------------------------------------
%
% The following footnote without marker is needed for the camera-ready
% version of the paper.
% Comment out the instructions (first text) and uncomment the 8 lines
% under "final paper" for your variant of English.
% 
\blfootnote{
    %
    % for review submission
    %
    % \hspace{-0.65cm}  % space normally used by the marker
    % Place licence statement here for the camera-ready version. See
    % Section~\ref{licence} of the instructions for preparing a
    % manuscript.
    %
    % % final paper: en-uk version 
    %
    % \hspace{-0.65cm}  % space normally used by the marker
    % This work is licensed under a Creative Commons 
    % Attribution 4.0 International Licence.
    % Licence details:
    % \url{http://creativecommons.org/licenses/by/4.0/}.
    % 
    % final paper: en-us version 

     \hspace{-0.65cm}  % space normally used by the marker
     This work is licensed under a Creative Commons 
     Attribution 4.0 International License.
     License details:
     \url{http://creativecommons.org/licenses/by/4.0/}.
     
}
The relationship between a group of human languages can be characterized across several dimensions of variation \citep{skirgaard2017some}, including (1) the temporal dimension, wherein languages have diverged from a common historical ancestor as in the case of Romance languages; (2) the spatial dimension, wherein the speaker communities are geographically adjacent as in the case of the Indo-Aryan and Dravidian languages of India; and (3) the socio-political dimension, wherein languages have evolved under shared political and/or religious forces as in the case of Arabic and Swahili. Languages, or language varieties, can be related across all these dimensions, which often results in a dialect continuum. Speakers of languages that constitute a dialect continuum can usually communicate with each other efficiently using their own mother tongue. The degree of intercomprehensibility between speakers of different language varieties within a continuum is mainly determined by linguistic similarities. A notable case of this phenomenon is the mutual intelligibility among the Slavic languages, which we study in this paper. \vspace{0.05cm}

One of the goals of linguistics is to study and categorize languages based on objective measures of linguistic distance. The degrees of similarity at different levels of the linguistic structural organization can be seen as preconditions for, as well as predictors of, successful oral intercomprehension.  For closely-related languages, similarities at the pre-lexical, that is the acoustic-phonetic and phonological, level have been found to be better predictors of cross-lingual speech intelligibility than lexical similarities \citep{gooskens2008phonetic, heeringa2009measuring}. In a different, yet relevant research direction, \cite{skirgaard2017some} have investigated non-linguists' perception of language variation using data from the popular spoken language guessing game, the Great Language Game (GLG). By analyzing the confusion patterns of the GLG's human participants, the authors have shown that factors predicting players' confusion in the game correspond to objective measures of similarity established by linguists. For example, both phylogenetic relatedness and overlap in phoneme inventories have been identified as factors of perceptual confusability (and by implication, similarity) of languages in GLG. \vspace{0.05cm}

The development of automatic systems that determine the identity of the language in a speech segment has received attention in the speech recognition community (see \cite{li2013spoken} for an overview). State-of-the-art approaches for automatic spoken language identification, henceforth LID, are based on multilayer deep neural networks (DNNs). DNN-based LID systems are parametric models that learn a mapping from spectral acoustic features of (untranscribed) speech to high-level feature representations in geometric space where languages are linearly separable. These models have shown tremendous success not only in discriminating between distant languages but also closely-related language varieties \citep{gelly2016language, shon2018convolutional, mateju2018using}. Nevertheless, none of the previous works in spoken language recognition has analyzed the emerging representations from neural LID models for related languages. Thus, it is still unknown whether the distances in these representation spaces correspond to objective measurements of linguistic similarity and/or to non-linguists' perception of language variation. In this paper, we aim to fill this gap and consider the family of Slavic languages as a case study. Our key contribution is two-fold: 
\begin{enumerate}[label={(\arabic*)}, noitemsep]
\item We present an LID model for Slavic languages based on convolutional neural networks. Our model incorporates a  domain-adversarial training strategy to improve its robustness against non-language sources of variability in speech signals. We show that our approach significantly improves generalization across datasets that differ in their recording conditions (\S\ref{sec:method} and \S \ref{sec:experiments}).  
\item We analyze and visualize the emergent representations from our robust LID model for 11 Slavic languages, five of which are not observed (held-out) during training. We show that the distances in the representation space correspond to measures of linguistic distance to a great extent (\S\ref{sec:analysis}).   \vspace{0.05cm}
\end{enumerate}
In this paper, we attempt to bridge different lines of research that have so far remained unconnected. On the one hand, we employ neural architectures from the field of spoken language recognition and build a robust model to identify languages in contemporary acoustic realizations of Slavic speech. On the other hand, we analyze the emerging language representations using techniques established by previous research in multilingual natural language processing (NLP). We consequently shed light on the speech modality and show how (untranscribed) speech signals can complement research done in computational studies of linguistic typology and language variation. 

% (comparative and historical) to the best of our knowledge

% The recognition of spoken language 

% LID in speech technology

% untranscribed speech 

% NN has made possible for end-to-end systems to be developed, while traditional approaches feature many components 

% closely-related languages have similar phonotactics, but differ in acoustic realizations of segments and suprasegmental features 

% language identity and objective linguistic measures of similarity 

% The GLG 

% similarity of representation in deep neural networks 

% --------------------------------------------------------------
\section{Background and Related Work}
% --------------------------------------------------------------

\subsection{Slavic Languages}
The Slavic language family is a branch of Indo-European languages that is conventionally divided into three subgroups: West-, East-, and South-Slavic. Apart from being related across the temporal dimension by sharing a common ancestor, Slavic languages form a spatial continuum of variation in a relatively connected geographic area across Europe and Northern Asia, except for the region where the Romance and Finno-Ugric wedge separates the South-Slavic from the West- and East-Slavic subgroups. Beside this traditional division (see Ethnologue, 23ed.), alternative classifications can be found in the Slavistics literature \citep[cf.][]{bednarczuk, pianka, dalweska2020, lehr, nalepa, manczak}.
% (cf. \cite{bednarczuk}, \cite{pianka}, \cite{dalweska2020}, \cite{lehr}, \cite{nalepa}, and \cite{manczak}).
% The Slavic language family is a branch of Indo-European languages with more than 300 million speakers. Apart from being related across the temporal dimension, that is; they share a common ancestor, Slavic languages form a spatial continuum since Slavic speakers today occupy a relatively unified geographic 
% The family of Slavic languages can be characterized across the temporal dimension by the common ancestor which in diachronic perspective leads to the Balto-Slavic continuum; as well as in the spatial dimension as relatively unified area with an exception of Romance and Finno-Ugric wedge separating South-Slavic from West- and East-Slavic subgroups.
% Apart from the most common division of the Slavic language family (e.g. Ethnologue, 23ed.), other groupings the were proposed (cf. \cite{bednarczuk}, \cite{pianka}, \cite{dalweska2020}, \cite{lehr}, \cite{nalepa}, and \cite{manczak}). 
Nevertheless, and despite differences in taxonomies among various proposals, the development of contemporary Slavic languages from a common historical ancestor is uncontroversial. The supporting arguments are based on historical phonology and comparative studies of the phoneme inventories \citep{sawicka}, as well as on studies of loanwords and Slavic toponyms. 
The high number of cognates as well as cross-linguistically shared features, such as lexical aspect, phonemic jotation and complex consonant clusters, provide strong evidence for common roots. In terms of diachronic phonology, the Common-Slavic era ends with the vocalization and reduction of the \textit{yers} -- the so-called ``half-vowels'' or ``reduced vowels'', [\begin{otherlanguage*}{russian}ъ\end{otherlanguage*}] and   [\begin{otherlanguage*}{russian}ь\end{otherlanguage*}]. The outcomes of these alternations consistently define the most common division of Slavic. Similarly, the reflexes of \textit{yat}  [\begin{otherlanguage*}{russian}ě\end{otherlanguage*}]  provide a clear distinction between East-, West- and South-Slavic. The results of common phonological processes, such as liquid metathesis, palatalization and sibilarization, also support the tripartite division. Moreover, these regularities of sound changes allow us to precisely trace the phonological development within the language family not only in the core standardized varieties but also in vernaculars and dialects. One of the objectives of our work is to assess the extent to which neural models of spoken language learn to detect such regularities from acoustic realizations of contemporary Slavic speech. 

% But regardless of differences in taxonomy, it is not questionable that contemporary Slavic languages arose from a common ancestor. The arguments for the unity came not only from the historical phonology and comparative studies of the phoneme inventories \citep{sawicka}, but also from the works on loanwords and Slavic toponyms. The high number of cognates and similarities between Slavic proper nouns suggest that division of the branch occurred relatively recently in comparison with other language families. In addition, the common features such as presence of lexical aspect, phonemic jotation, and complex consonant clusters, provide a strong argument for the common Slavic roots.

% which appeared as a consequence of dissimilation of the yers, 

% Hence, the above-mentioned similarities create a challenge for our task in a phonetic and phonotactic perspective. 

%^^^044A %[ě]

\subsection{Language Identification in Speech Signals}
Research in automatic identification of the language in a speech signal  \citep{li2013spoken} is mainly concerned with the development of computational models that take  an acoustic realization of a short utterance (usually a few seconds of speech) and predict the spoken language as output. Currently, end-to-end deep neural networks are the predominant paradigm for LID and have shown tremendous success in previous works \citep{lopez2014automatic, gonzalez2014automatic, gelly2017spoken}. In this paradigm, the LID problem has usually been modelled as a temporal sequence classification problem in which a spectro-temporal representation of a spoken utterance (e.g., a sequence of spectral feature vectors) is transformed via a multi-layer neural network into a high-level vector representation that captures language-ID features. Other works in the literature have addressed LID for closely-related spoken language varieties including Arabic dialects \citep{shon2018convolutional, gelly2016language}, Iberian languages \citep{gelly2016language}, and Slavic languages \citep{mateju2018using, Abdullah2020}. 

At the intersection of speech recognition and linguistic typology, \cite{gutkinpredicting} have trained a neural network on a large-scale multilingual speech database to predict typological features of the World Atlas of Language Structure (WALS)  \citep{wals} for a language given a speech segment.  The authors have shown that the speech modality contains enough signals to predict typological features of a held-out set of languages without explicit linguistic annotations. Their findings indicate that neural networks trained on multilingual speech could capture linguistic regularities and generalize beyond the languages observed in the training data.  %However, we are not aware of any previous work that has analyzed the emerging representations from LID models or investigated whether or not the distance in these representation spaces reflect the linguistic distance. 

% of the speech signal into a high-level feature vector 
% Spoken language identification (LID) are suitable candidates for the task of quantifying acoustic similarity. LID models take as an input an acoustic realization of a linguistic expression (usually a few seconds of a spoken utterance) and produce as an output a probability distribution over the candidate languages.

% mention research that did closely-related language LID 

% what are the key findings of the previous work in this line of research

% mention typological feature prediction from speech paper

\subsection{Language Representations in Continuous Vector Spaces}

Inspired by the advances in representation learning for NLP, multilingual neural models have been explored in the literature to induce real-valued language vectors, also known as \textit{language representations} or \textit{language embeddings}, where a single vector ($\mathbf{v} \in \mathbb{R}^d$) is associated with each language. Even though it has been motivated from different points of view, the main idea of this stream of research is to train a single NLP model on many languages whereby the language representation space is learned by exploiting the multilingual signal. For example, \citet{johnson2017google} introduced a multilingual neural machine translation (NMT) model in which the required target language of the translation was specified by the language embedding. Other works have either scaled this approach to a massively multilingual setting \citep{ostling-tiedemann-2017-continuous, malaviya-etal-2017-learning} or explored other NLP tasks such as linguistic structure prediction \citep{bjerva2019language} and grapheme-to-phoneme conversion \citep{peters-etal-2017-massively}. Furthermore, \cite{rabinovich-etal-2017-found} and \cite{bjerva2019language} have analyzed  the learned language representations and shown  that the distance in the  representation space reflects the phylogenetic distance between Indo-European languages. However, \cite{bjerva2019language}  have argued that structural syntactic similarities between languages are a better predictor of the language representation similarities  than phylogenetic similarities. 

%, which is then transformed into a language embedding, at the beginning of each input sentence
The most relevant analysis to ours is the recent work by \cite{cathcart-wandl-2020-search}, in which the authors have trained a neural sequence-to-sequence model on a Slavic etymological dictionary. Their model was trained to consume a reconstructed Proto-Slavic word form and a language embedding, then emit a word form in the modern language specified by the language embedding. The authors have applied a clustering analysis on the learned language embeddings and successfully reconstructed the phylogenetic Slavic family tree. Our work complements this line of research with one fundamental difference: we perform our analysis on contemporary realizations of Slavic speech instead of the historically reconstructed phonological data without explicitly training our model to capture systematic sound changes. 

% that their models was able to learn diachronic  phonological  generalizations and they were able , The contemporary word forms  are represented as a sequence of phonemes in IPA transcriptions.

% that how they can be used to measure similarities between languages and

%\subsection{}

% --------------------------------------------------------------
\section{Methodology}
% --------------------------------------------------------------
\label{sec:method}

\subsection{Slavic Speech Data}
\label{sec:slavic_data}
The data we use in this research are drawn from two different datasets:

\begin{enumerate}[label={(\arabic*)}, noitemsep]
\item  \textbf{Radio Broadcast Speech (RBS)} \hspace{0.1cm} A large collection of Slavic speech recordings were collected by crawling online radio stations in previous work \citep{nouza2016asr, mateju2018using}. The dataset includes speech segments in 11 Slavic languages from the three subgroups: (1) South-Slavic: Bulgarian (\textsc{bul}), Croatian (\textsc{hrv}), Serbian (\textsc{srp}), Slovene (\textsc{slv}), and Macedonian (\textsc{mac}). (2) West-Slavic: Czech (\textsc{cze}), Polish (\textsc{pol}), and Slovak (\textsc{slo}). (3) East-Slavic: Russian (\textsc{rus}), Ukrainian (\textsc{ukr}), and Belorussian (\textsc{bel}). The audio recordings are either segments of professional news reports or of spontaneous speech during discussions.  The recording conditions are diverse and the utterances occasionally include background music. We sample 8,000 and 500 utterances per language from the training split as our training and validation sets, respectively, and use the test set in \cite{mateju2018using} as our evaluation set. 

% This dataset does not include any speaker metadata. Thus, we are not certain whether or not the training and evaluation speakers in this dataset are disjoint.  

\vspace{0.15cm}

\item \textbf{GlobalPhone Read Speech (GRS)} \hspace{0.1cm} We also use the Slavic portion of the multilingual GlobalPhone speech database \citep{schultz2013globalphone} which includes  read speech recordings from native speakers of six Slavic languages: Bulgarian, Croatian, Czech, Polish, Russian, and Ukrainian. The utterances vary in length and quality across languages. Our final GRS training subset consists of 8,000 utterances per language. 

% We set the minimum utterance length to 3 seconds and segment longer utterances into non-overlapping 3-second speech segments. 

% We use the same splits as in \cite{9053144}. \vspace{0.15cm} 
\end{enumerate}

\subsection{Signal Representations of Speech}
Human speech can be modelled with various signal representations. For automatic speech recognition (ASR), the conventional approach is to convert time-varying speech waveforms into time-frequency, or spectro-temporal, representations using a standard signal processing pipeline based on the short-time Fourier transform. An example of such representations is the mel-frequency spectral coefficients (MFSCs) representation, whose development has been inspired by the human auditory system.  MFSCs describe the spectral envelope along the temporal dimension in a way that reflects the shape of the human vocal tract during speech production at each timepoint. In this paper, we use MFSCs for all presented experiments.

%Previous work usually refers to MFSCs as filterbanks \citep{shon2018convolutional}, but we choose to follow \citep{mohameddeep} in using the term MFSCs to refer to the mel-frequency spectral features that are correlated. 

%MFCCs acoustic vectors as proxy of acoustic features 

\subsection{Neural LID Model}
\label{sec:neural_model}
%We define the LID task as a discriminative sequence classification problem. First, a variable-length speech segment is transformed by an acoustic front-end into a sequence of acoustic observations $\mathbf{X} = (\mathbf{x}_1, \dots, \mathbf{x}_T)$, where $\mathbf{x}_t \in \mathbb{R}^k $ is a spectral-based feature vector  at timestep $t$. Given a  sequence  $\mathbf{X}$, the goal is to predict the spoken language $\hat{y}$. Using a deep neural network as a classification model, the LID problem can be defined as 

Our problem definition and LID models are based on the work of \cite{Abdullah2020}. The LID task is defined as an instance of temporal sequence classification. A speech segment is first converted into a sequence of acoustic events $\mathbf{X} = (\mathbf{x}_1, \dots, \mathbf{x}_T)$, where $\mathbf{x}_t \in \mathbb{R}^k $ is a spectral feature vector  at timestep $t$. Then, the goal is to predict the spoken language $\hat{y}$ given the sequence $\mathbf{X}$. This definition can be formalized using a deep neural network as a parameterization of the model as follows 
\[ \hat{y} =  \underset{y \in \mathcal{Y} }{\arg\max} \: P(y \; | \; \mathbf{X};\;  \boldsymbol{\theta})  \]
% \begin{equation} \end{equation}
where $\mathcal{Y}$ is the set of languages, $\boldsymbol{\theta}$ is the model's parameters learned from a labelled dataset, and  $P(y  |  \mathbf{X}; \boldsymbol{\theta})$ is the posterior probability of the language label $y$. 
\vspace{0.1cm}

\begin{figure}[t]
  \begin{minipage}[c]{0.54\textwidth}
    \includegraphics[width=\textwidth]{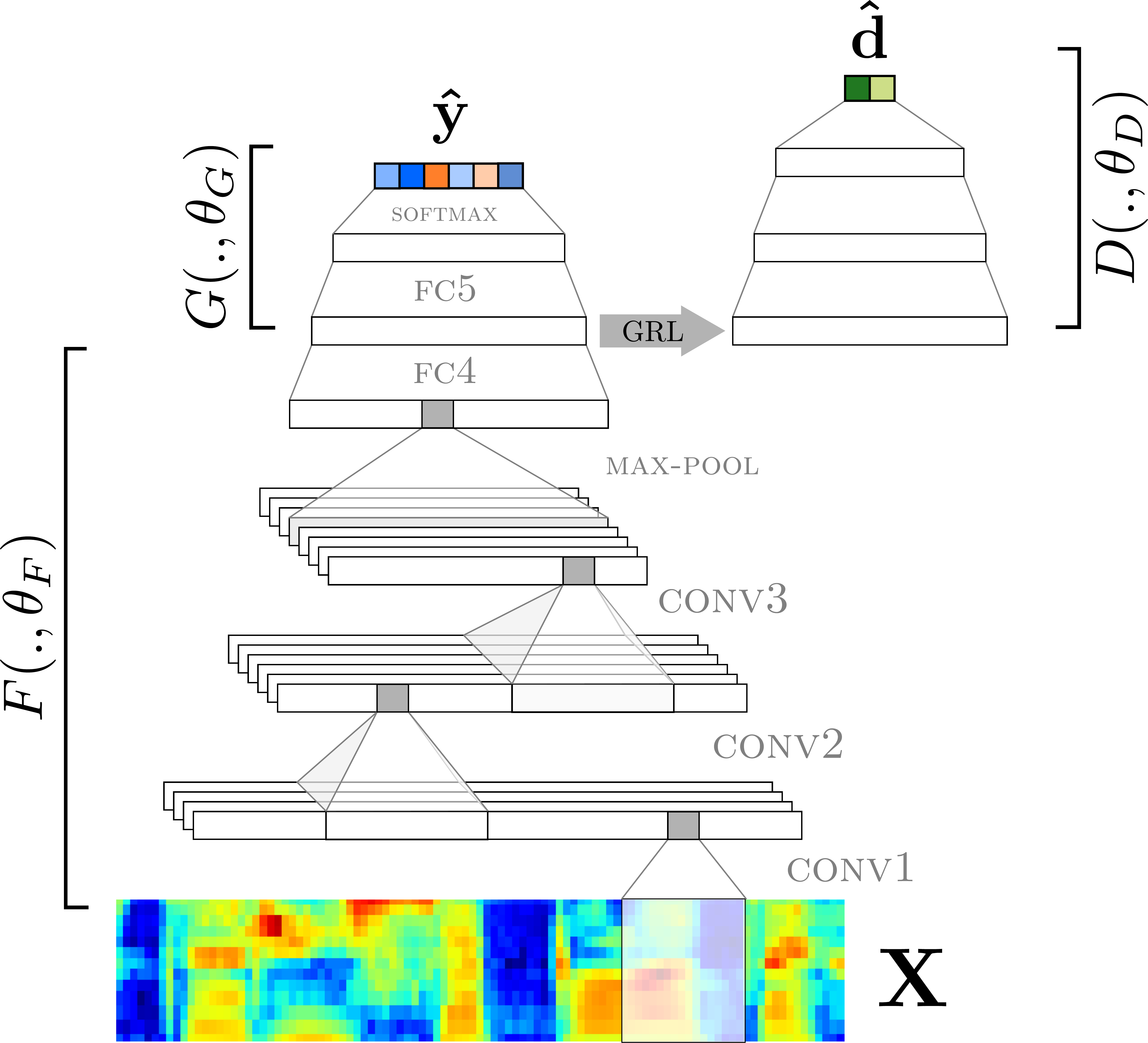}
  \end{minipage}\hfill
  \begin{minipage}[c]{0.42\textwidth}
    \caption{A schematic view of our LID model: a 3-layer 1-dimensional convolutional network followed by 3-layer fully-connected feed-forward network. The segment-level feature extractor $F(., \boldsymbol{\theta}_F)$ consists of layers \textsc{conv1}, \textsc{conv2}, \textsc{conv3}, \textsc{max-pool}, and \textsc{fc4}, and maps the input sequence $\mathbf{X}$ into a $d$-dimensional feature vector $\mathbf{f} \in  \mathbb{R}^d$, i.e. $\mathbf{f} = F(\mathbf{X}; \boldsymbol{\theta}_F)$. Then, the language classifier $G(.,\boldsymbol{\theta}_G)$, which consists of layers \textsc{fc5} and \textsc{softmax}, maps $\mathbf{f}$ into a probability distribution over the language space, i.e. $\mathbf{\hat{y}} = G(\mathbf{f}; \boldsymbol{\theta}_G)$. In the robust LID model, the domain classifier $D(.,\boldsymbol{\theta}_D)$ is connected to the network through a gradient reversal layer (GRL) to predict the domain given $\mathbf{f}$. The GRL behaves as an identity function during the forward pass but reverses the direction of the gradient signal during backpropagation.
    } 
    \label{fig:nn}
  \end{minipage}
\end{figure}

\noindent
\textbf{Baseline.}  \hspace{0.25cm} We use an end-to-end convolutional neural network (CNN) with three convolutional layers followed by three feed-forward layers (see Fig. \ref{fig:nn} for full description of the model). Our baseline LID model can be viewed as two components that are jointly trained: a \textit{segment-level feature extractor} ($F$) and a \textit{language classifier} ($G$), each associated with two sets of parameters  $\boldsymbol{\theta}_F$ and $\boldsymbol{\theta}_G$,  respectively. The parameters of the network $\boldsymbol{\theta}_F$ and $\boldsymbol{\theta}_G$ are learned in a supervised approach given a source dataset ${\mathcal{D}_\mathcal{S}} = \{(\mathbf{X}_i, y_i)\}_{i=1}^{N_\mathcal{S}}$ of ${N_\mathcal{S}}$ labelled samples and an optimization algorithm that minimizes cross-entropy loss.  

% We consider our LID model to be two components  The parameters of the network $\boldsymbol{\theta}_f$ and $\boldsymbol{\theta}_y$ are learned given a dataset $\mathcal{D}_\mathcal{S} = \{(\mathbf{X}_{i}, y_{i})\}_{i=1}^{N_\mathcal{S}}$ of ${N_\mathcal{S}}$ labelled samples by minimizing the objective function 
% %a large set of labelled training examples in one domain, i.e., $\mathcal{S} = \{(\mathbf{X}_1, y_1), \dots, (\mathbf{X}_N, y_N)\}$.  
% \begin{equation}J(\boldsymbol{\theta}_f, \boldsymbol{\theta}_y) = \sum_{(\mathbf{X}_i, y_i) \in \mathcal{D}_\mathcal{S}}^{}L_y\Big(G_y\big(G_f(\mathbf{X}_i; \boldsymbol{\theta}_f); \boldsymbol{\theta}_y\big), y_i\Big) \end{equation}
% where $L_y$ is the loss of the language classifier.  

% \vspace{-0.5cm}

\noindent
\textbf{Robust LID.}  \hspace{0.25cm} To improve the robustness of our model against non-linguistic sources of variability in speech signals, we employ a well-established adversarial domain adaptation strategy  \citep{ganin2015unsupervised}, which has been shown to be effective for LID \citep{Abdullah2020}. Adversarial domain adaptation aims to minimize the discrepancy between the representations of the model given speech samples from two sources that differ in their recording conditions or spoken genre. This approach only requires a target dataset ${\mathcal{D}_\mathcal{T}} = \{(\mathbf{X}_i)\}_{i=1}^{N_\mathcal{T}}$ of ${N_\mathcal{T}}$  unlabelled samples, in addition to the labelled samples of the source dataset $\mathcal{D}_\mathcal{S}$ to train the model.  To this end, an adversarial \textit{domain classifier} ($D$) with  parameters $\boldsymbol{\theta}_D$ is connected to the network with the objective of predicting the dataset of each speech sample in the training data. Now the goal is to encourage the segment-level feature extractor $F$ is to produce representations that are language-discriminative but invariant with respect to the non-linguistic sources of variability in speech signals.  We refer the reader to previous work to get a detailed overview of adversarial training for domain adaptation in the context of the  LID task \citep{Abdullah2020}. 

% The overall objective function is to minimize 
% \begin{multline}
% J(\boldsymbol{\theta}_f, \boldsymbol{\theta}_y, \boldsymbol{\theta}_d) =  \sum_{(\mathbf{X}_i, y_i) \in \mathcal{D}_s}^{}L_y\Big(G_y\big(G_f(\mathbf{X}_i; \boldsymbol{\theta}_f); \boldsymbol{\theta}_y\big), y_i\Big) \\ - \lambda \sum_{(\mathbf{X}_i, d_i) \in (\mathcal{D}_s \cup \mathcal{D}_t)}^{} L_d\Big(G_d\big(G_f(\mathbf{X}_i; \boldsymbol{\theta}_f); \boldsymbol{\theta}_d\big), d_i\Big)
% \end{multline}
% where $L_y$ is the loss of the language classifier, $L_d$ is the loss of the domain classifier, and $\lambda$ is a parameter that controls the contribution of the domain classifier's loss to the overall loss. In practice, this adversarial loss is realized with a special layer that reverses the direction of the gradient signal coming from the domain classifier's loss into the feature extractor during backpropagation, which is referred to as a gradient reversal layer. This procedure is an instance of adversarial learning where different blocks in the network are trained with competing objectives.

% --------------------------------------------------------------
\section{Experiments and Results}
% --------------------------------------------------------------
\label{sec:experiments}

\subsection{\textbf{Training and Evaluation Data}}
We train our LID models to discriminate between the six Slavic languages that are shared by the RBS and GRS datasets, namely: Bulgarian, Czech, Croatian, Polish, Russian, and Ukrainian. The labelled speech samples are drawn from the RBS dataset (source dataset) while the unlabelled speech samples, which are used to improve our model robustness against non-language sources of variability, are drawn from the the GRS dataset (target dataset). For each language, we sample a balanced subset of 3-second 8,000 segments from each dataset to ensure our models are not effected by undesirable biases due to imbalanced conditions. We evaluate our models in two conditions: (1) in-domain evaluation, in which the evaluation samples come from the source dataset (i.e., the test split of RBS), and (2)  cross-domain evaluation, in which the evaluation samples come from the target dataset whose labels are not observed during training (i.e., the development split of GRS).

% The GRS dataset includes only a subset of languages from the RBS dataset, namely  Bulgarian, Czech, Croatian, Polish, Russian, and Ukrainian. For the analysis in this paper, we train our models using the 6 Slavic languages that are shared by the two datasets. Our models are trained on the labelled speech samples from the RBS and the speech samples from the GRS dataset are only used to encourage the network to learn domain-invariant representations for the LID task. \banote{If the paper to InterSpeech gets accepted, a 'self-citation' will be added here.}

\noindent
\subsection{Feature Extraction}
In our experiments, we use the first 12 mel-frequency spectral coefficients (MFSCs) and frame-level averaged energy as low-level speech features.  We extract frames of 25ms with 10ms overlap. Then, each speech sample is normalized with utterance-level mean and variance normalization.

\noindent
\subsection{Model Architecture and Hyperparameters}
\textbf{CNN Architecture.} \hspace{0.1cm} We employ three 1-dimensional convolutional layers over the temporal dimension with 128, 256, and 512 filters and filter widths of 5, 10, and 10 for each layer with strides of 1 step for each layer. Batch normalization and ReLU non-linearity are applied after each convolutional operation. We downsample the representation by applying a single max pooling operation at the end of the convolution block. For the language classifier, we use three fully-connected layers (512 $\rightarrow$ 512 $\rightarrow$ 512 $\rightarrow$ 6) before the softmax for both the baseline and the robust LID models. \vspace{0.1cm}

\noindent
 \textbf{Adversarial Classifier.} \hspace{0.1cm} For our robust LID model, we add a 3-layer fully-connected feed-forward block (512 $\rightarrow$ 1024 $\rightarrow$ 1024 $\rightarrow$ 2) to the network as the adversarial domain classifier $D$ .  The adversarial classifier takes the output of the segment-level feature extractor $\mathbf{f}$ (the output of layer \textsc{fc4}) and predicts the domain of the input sample. The adversarial loss is realized with a special layer, that is, a gradient reversal layer (GRL). \vspace{0.1cm}

% We consider the feature extractor as the convolutional block as well as the first layer of the fully-connected block; thus, the reversed gradient signal from the adversarial classifier is back-propagated into all layers of the network except the final layer before the softmax of the language classifier. \vspace{0.1cm}

% For the adaptation factor $\lambda$, we use a gradually increasing value $\in [0, 1]$ to suppress the noise from the feature extractor during the initial phase of the training procedure.

\noindent
\textbf{Training Details.} \hspace{0.1cm} The cross-entropy loss is used for both the language classifier loss and the adversarial classifier loss. We use the Adam optimizer with a learning rate of $1 \times 10^{-3}$ and train our models with a batch size of 256 samples. For the robust model, half of the samples within a mini-batch are drawn from the training split of the RBS dataset while the rest are drawn from the training split of the GRS dataset. Both the baseline and the robust models are trained for 50 epochs and the best models are selected based on the performance on the validation set. We do not use the early stopping criterion in our experiments. \vspace{0.1cm}

\subsection{Experimental Results}
\label{sec:results}

%Baseline model is trained without adversarial classifier, while the AT+ model employs an adversarial classifier as an unsupervised domain adaptation.
% We evaluate our model in two conditions: (1) in-domain evaluation, whereby the evaluation samples come from the same training dataset (i.e., the test split of RBS), and (2)  cross-domain evaluation, whereby the evaluation samples come from the other dataset whose labels are not observed during training (i.e., the development split of GRS).

In this section, we report the evaluation results and show the effect of the adversarial training on the model's robustness.  Since the GRS evaluation data is imbalanced, we use balanced accuracy \citep{brodersen2010balanced} as our evaluation metric to obtain a better estimate of the models' performance. Table \ref{tab:main_results} shows our results for speech segments of various lengths. It can be observed from the cross-domain evaluation that the performance of the baseline model drops by a substantial factor. On the other hand, our robust model significantly improves cross-domain performance with little effect on the performance on the in-domain evaluation dataset, especially for longer utterances. 

To get further insight into why adversarial training improves cross-domain performance, we analyze the predictions of the GRS evaluation samples made by the baseline and robust models by computing the $F_1$ score per language. The results of this analysis are shown in Table \ref{tab:f_score}. For our baseline, we observe a much higher variance between languages compared to the robust model. The performance drop is more pronounced in the case of Ukrainian with a significant decrease in $F_1$. However, our robust model boosts the $F_1$ score on Ukrainian from 14.66\% to 94.49\%.%, which is the highest among other languages.

\begin{table}[t]
\centering
\begin{tabular}{@{}rccccccccc@{}}
\toprule
       & \multicolumn{4}{c}{\textbf{In-domain}} &  & \multicolumn{4}{c}{\textbf{Cross-domain}} \\ \cmidrule(lr){2-5} \cmidrule(l){7-10} 
\multicolumn{1}{c}{\textbf{LID Model}}      & 1-sec    & 2-sec    & 3-sec    & Full     &  & 1-sec     & 2-sec     & 3-sec     & Full     \\ \midrule
 Baseline LID  & 72.93    & 91.10    & 95.48    & 97.38    &  & 30.18     & 47.61     & 55.91     & 65.45    \\
Robust LID  & 64.25    & 88.55    & 94.77    & 97.35    &  & 51.59     & 76.76     & 86.94     & 93.29    \\ \midrule
$\Delta$       &   -11.9        &  -2.80        &   -0.74       &    -0.03      &   &  70.94    &   61.23     &    55.50    &  42.54        \\ \bottomrule
\end{tabular}
\caption{In-domain and cross-domain evaluation of our LID models in balanced accuracy (\%). $\Delta$ is the relative percentage difference in accuracy scores of baseline and robust LID models. }
\label{tab:main_results}

\end{table}

\vspace{2pt}
\vspace{0.5cm}

\begin{table}[h]
\centering
\begin{tabular}{@{}rcccccccccc@{}}
\toprule
\multicolumn{1}{l}{\textbf{LID Model}} &  & \textsc{bul}   & \textsc{hrv}   & \textsc{cze}   & \textsc{pol}   & \textsc{rus}   & \textsc{ukr}   &  & macro Avg.  & micro Avg.  \\ \midrule
Baseline LID                     &  & 59.73 & 64.86 & 76.50 & 61.93 & 41.79 & 14.66 &  & 53.25 & 54.17 \\
Robust LID                           &  & 85.12 & 83.32 & 89.36 & 83.96 & 87.66 & 94.49 &  & 87.32 & 88.26 \\ \midrule
$\Delta$                               &  & 42.51     &  28.46      &  16.81     &     35.57  &   109.76     &    544.54    & &  63.98     &   62.93    \\ \bottomrule
\end{tabular}
\caption{$F_1$ score (\%) per language. Predictions were obtained by feeding 3-second segments from GRS evaluation dataset to the LID models. $\Delta$ is the relative percentage difference in $F_1$ scores of baseline and robust LID models.}
\label{tab:f_score}

\end{table}

\vspace{1cm}

% For example, while the baseline model achieves up to 76.50\% $F_1$ score on Czech, it drops to 14.66\% on Ukrainian, which is worse than chance-level performance (16.7\%). This drop in performance for the case of Ukrainian is most likely due to discrepancy in the recording conditions between the two datasets we use in our study. The advantage of adversarial training is demonstrated in Table \ref{tab:f_score}.

% \subsection{LID Performance}

% --------------------------------------------------------------
\section{Representation Similarity Analysis}
% --------------------------------------------------------------
\label{sec:analysis}

In this section, we present our representation similarity analysis. From the robust LID model presented in the previous section, we obtain 512-dimensional representations for the 11 languages from the RBS evaluation set (500 5-second speech segments for each language). These representations are the output of the last fully-connected layer of the language classifier (i.e., layer \textsc{fc5}) before the softmax output layer.\footnote{In the speaker and language recognition community, these representations are usually referred to as $\mathbf{x}$-vectors, while the emerging geometric space is referred to as the $\mathbf{x}$-vector space.} It is worth pointing out that our LID model has been trained on a subset of only six languages out of the 11 languages that we analyze in this section.

\subsection{Language Representation Visualizations}

\begin{figure}[t]
     \centering
     \begin{subfigure}[b]{0.45\textwidth}
         \centering
         \includegraphics[width=\textwidth]{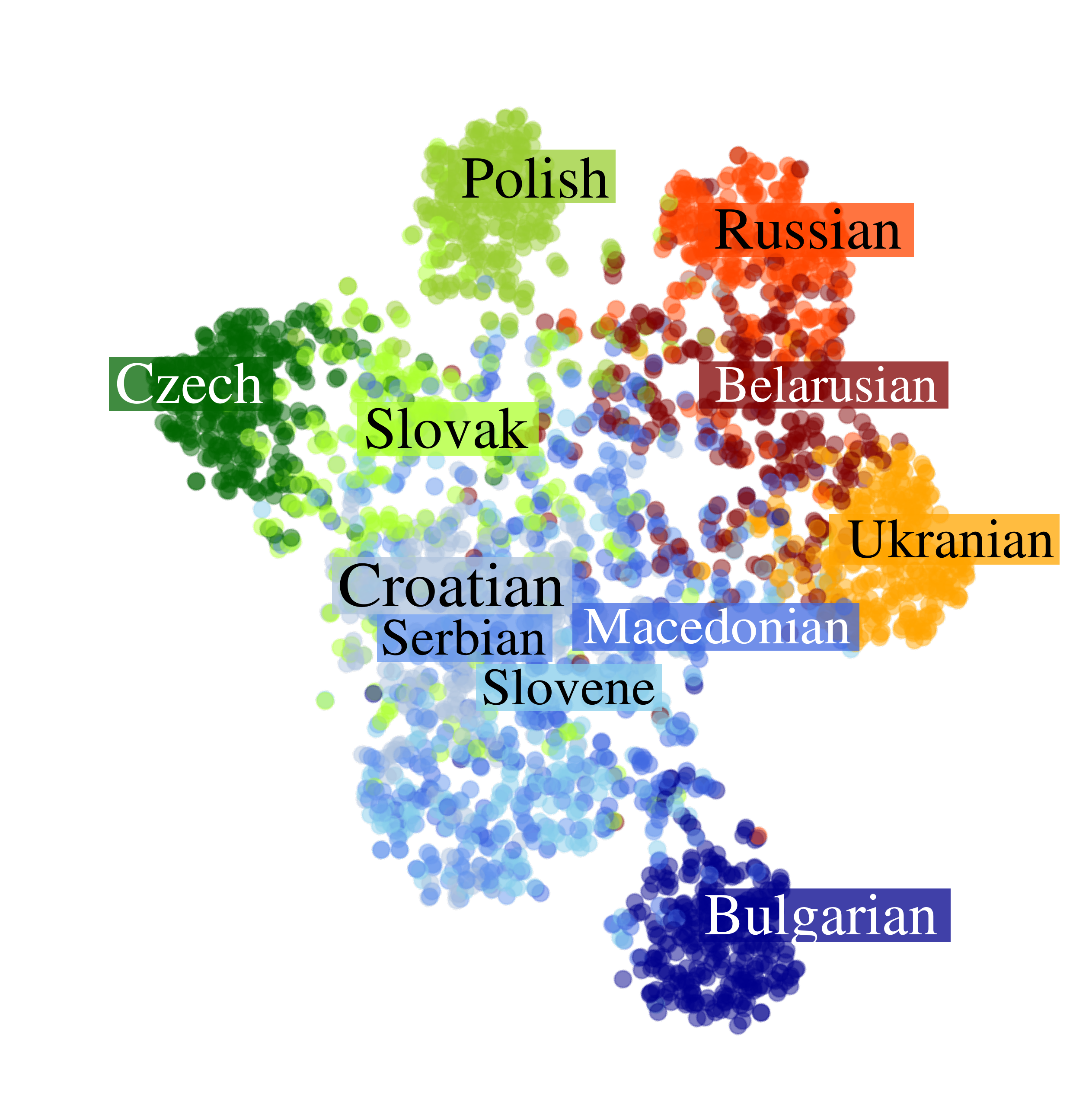}
         \caption{}
         \label{}
     \end{subfigure}
     \hfill
     \begin{subfigure}[b]{0.45\textwidth}
         \centering
         \includegraphics[width=\textwidth]{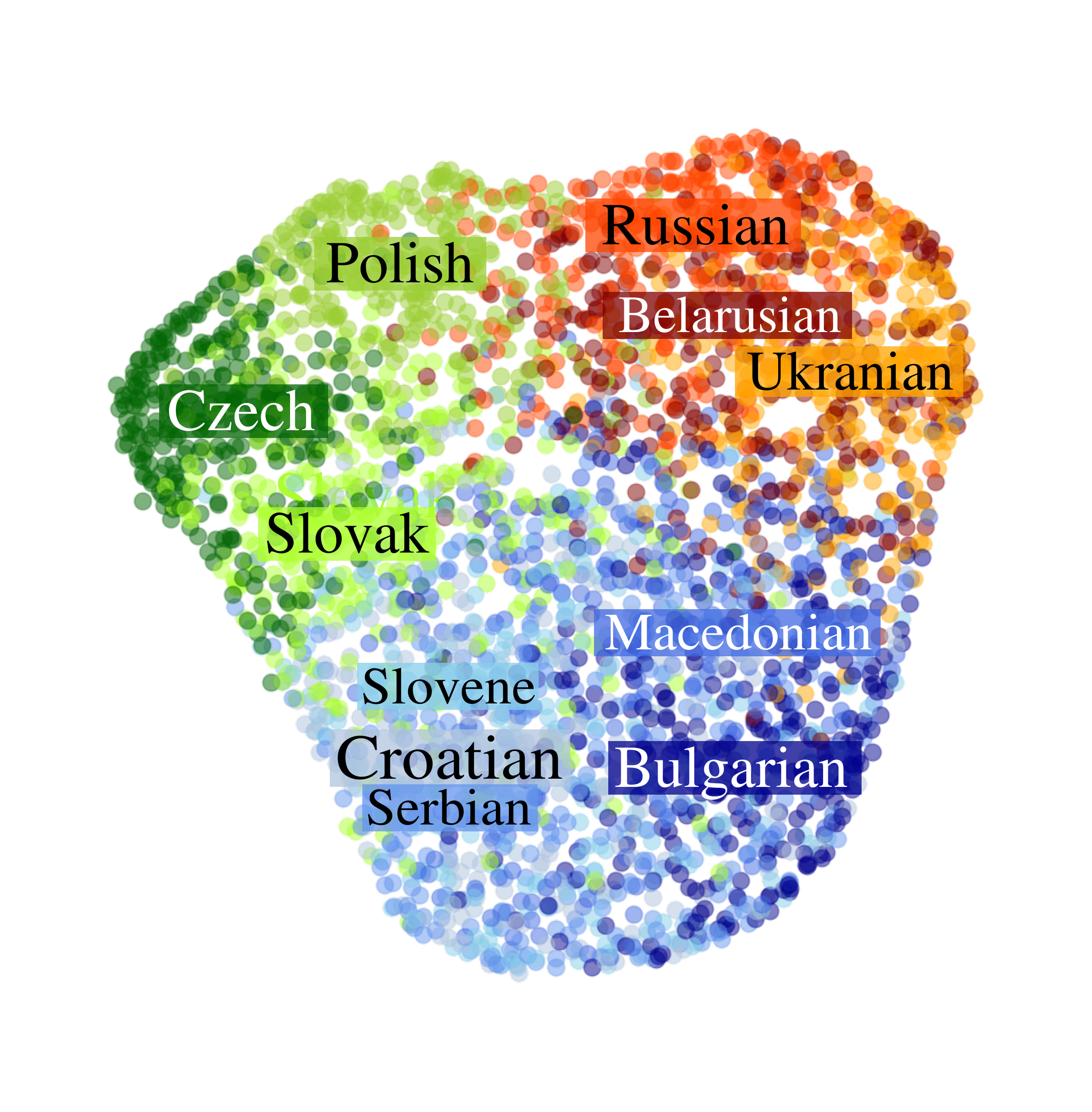}
         \caption{}
         \label{}
     \end{subfigure}
     \vspace{25pt} 
     \caption{Two-dimensional visualization of representations of evaluation speech segments: (a) t-SNE projections, and (b) UMAP projections (best viewed in color).}
    \label{fig:viz} 
\end{figure}

As our first analysis, we use dimensionality reduction techniques to obtain 2-dimensional projections from representations of the evaluation set and visualize the resulting data points. We use two dimensionality reduction techniques; t-SNE \citep{maaten2008visualizing} and UMAP \citep{mcinnes2018umap}. The resulting graphs are illustrated in Fig. \ref{fig:viz}. The motivation for using two different techniques in this analysis is that t-SNE and UMAP have different optimization objectives that complement each other. That is, the t-SNE algorithm preserves the local structure of the space; thus, it mainly reveals the cluster structure within the representation space. On the other hand, the UMAP algorithm  preserves the global structure of the space. Nevertheless, both t-SNE and UMPA plots in Fig. \ref{fig:viz} show very similar trends since the emerging subspaces shown in the figure correspond to the conventional sub-grouping of Slavic languages into East-, West-, and South-Slavic.

% Nevertheless, the t-SNE plot in Fig. \ref{fig:viz}(a) and the UMAP plot in  Fig. \ref{fig:viz}(b) show very similar trends, as the regions in the figures reflect to a large extent with the traditional grouping of Slavic languages into East, West, and South Slavic sub-groups. 

\subsection{Correlation with Geographic Proximity}
Following \cite{bjerva2019language}, we investigate whether the language representations reflect the geographic proximity of their respective speaker communities.  To perform this analysis, we first obtain a single prototypical  vector representation for each Slavic language in our study by taking the average over the representations of the evaluation speech segments as 

\[\mathbf{v}_L =\frac{1}{|\mathcal{E}_L|} \sum_{\mathbf{X} \in \mathcal{E}_L}\texttt{NN}(\mathbf{X}) \] 

\noindent
where $\mathbf{v}_L \in \mathbb{R}^{512}$ is a prototypical vector representation for language $L$, $\mathcal{E}_L$ is the evaluation speech segments for language $L$, and $\texttt{NN}(\mathbf{.})$ is the output of the last non-linear layer (layer \textsc{fc5}) of the robust LID model. The distance in the representation space between two languages is computed using cosine distance. For geographic distance, we follow a similar approach as in \cite{skirgaard2017some}. First, each language is characterized by a point location on the map given the latitude and longitude information in the ASJP linguistic database (which is intended to represent the cultural or historical center of the language).  We then compute the pairwise distances between the points on the map (in kilometers) and convert them into log$_{10}$ scale. Fig. \ref{fig:corr}(a) shows a scatter plot between the data points in which the $x$-axis represents the geographic distance and the $y$-axis represents the cosine distance in the representation space. We observe a positive correlation between the two distance measures (Pearson's $r =  0.58$). This clearly shows that the distance in the representation space does indeed reflect the geographic distance.

% as follows

% \[  d(\mathbf{v}_A,  \mathbf{v}_B) =  1 - \frac{\mathbf{v}_A  \,  \mathbf{\cdot} \,  \mathbf{v}_B}{\norm{\mathbf{v}_A} \norm{\mathbf{v}_B}}  \] 

% \noindent

\subsection{Genetic Signal in the Representation Space}
\begin{figure}[t]
     \centering
     \begin{subfigure}[b]{0.48\textwidth}
         \centering
         \includegraphics[width=\textwidth]{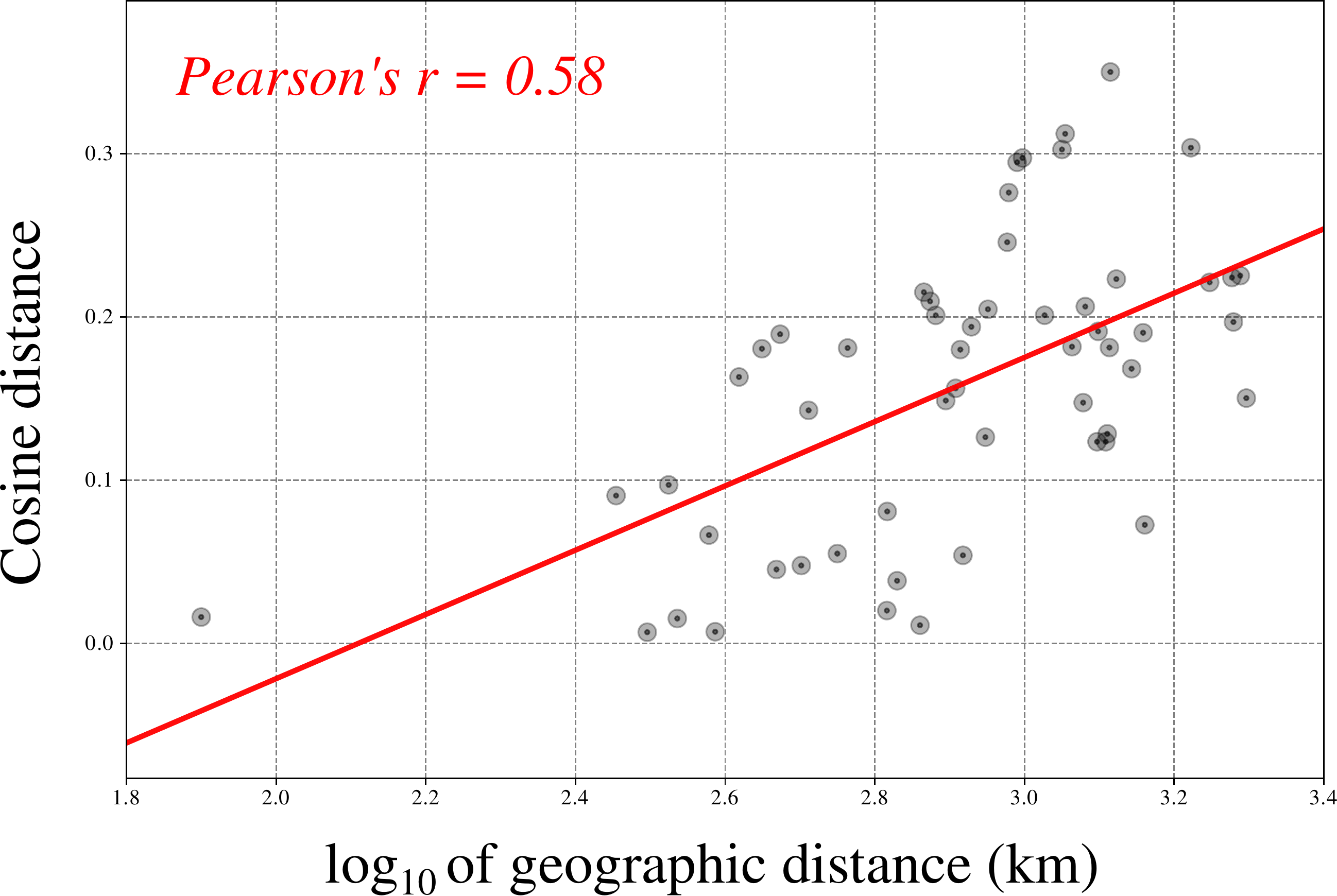}
         \caption{}
         \label{}
     \end{subfigure}
     \hfill
     \begin{subfigure}[b]{0.42\textwidth}
         \centering
         \includegraphics[width=\textwidth]{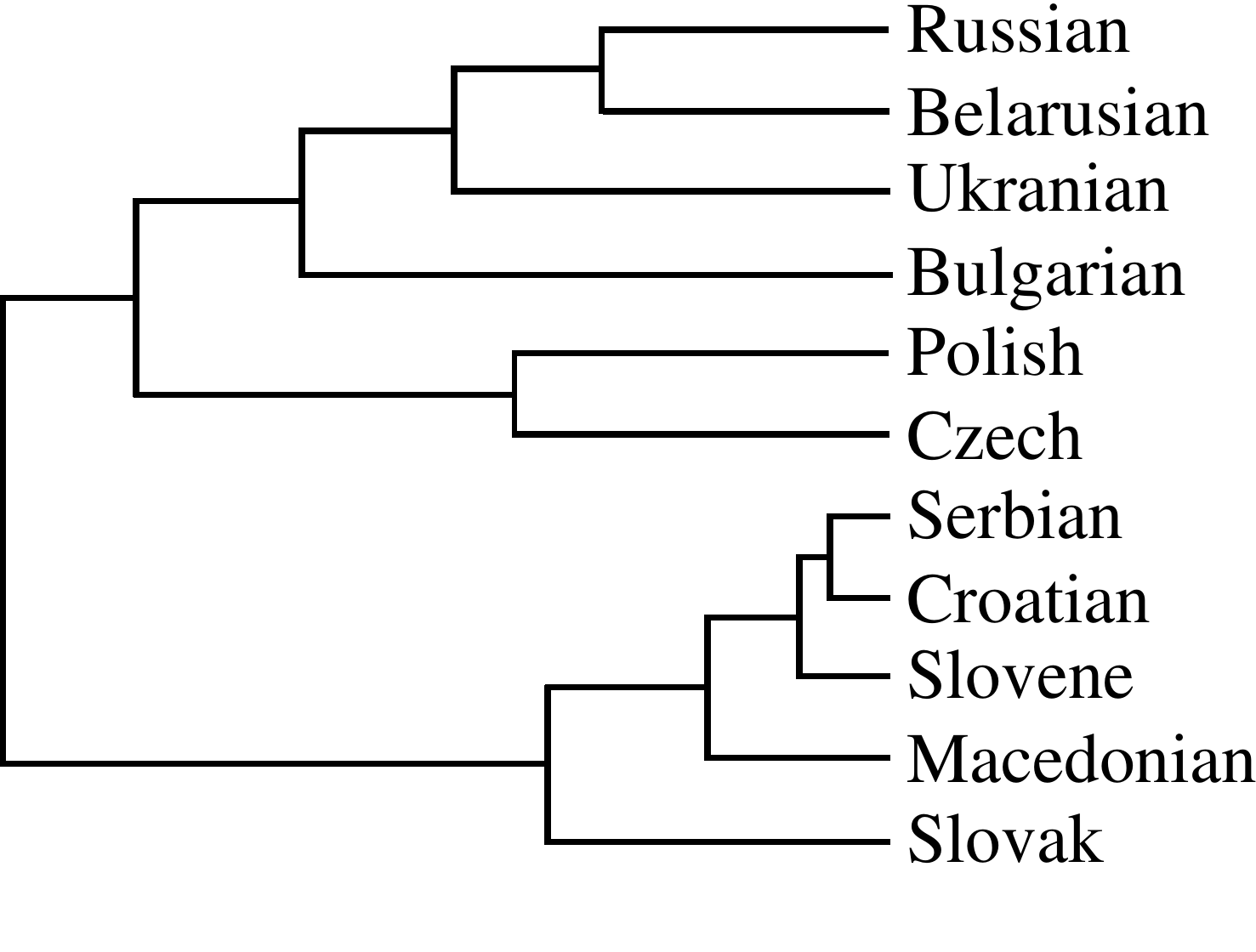}
         \caption{}
         \label{}
     \end{subfigure}
      \vspace{25pt} 
     \caption{(a) Correlation between geographic distance and distances between prototype language representations measured by cosine similarity, and (b) a genetic tree generated from the pairwise distance matrix of the language prototype representations from our LID model.}
     \label{fig:corr}
\end{figure}

Similar to the analysis in  \cite{bjerva2019language} and \cite{cathcart-wandl-2020-search}, we investigate the genetic signal in the representation space. To this end, the pairwise cosine distances computed in the previous section are first converted into a confusion matrix. Then, we generate a tree by performing hierarchical clustering on the confusion matrix using the Ward algorithm. The resulting tree is depicted in Fig. \ref{fig:corr}(b). We observe that the generated tree shows many similarities to phylogenetic trees that correspond to the widely accepted tripartite division of Slavic languages. 

% There are two exceptions that do not agree with the traditional wisdom of Slavic taxonomy: (1) Bulgarian is grouped with East-Slavic languages, and not South-Slavic subgroup, and (2) Slovak is grouped with with the South-Slavic languages, and not the West-Slavic subgroup. 

\subsection{Fine-grained Analysis of Similarities}
Although we have shown that the distances in the representation space of our LID model correlate with geographic distances and have adequately reconstructed the Slavic genetic tree, we seek to understand the factors that could better explain the similarities.  To this end, we compare our generated tree with various genetic trees of Slavic languages presented in previous studies.\footnote{Although a correlation analysis on confusion matrices would have been more optimal, we instead apply tree similarity analysis because not all underlying confusion matrices are available.  }

\begin{enumerate}[label={(\arabic*)}, noitemsep]
\item \textbf{Levenshtein distance-based tree} \hspace{0.1cm}  \cite{serva2008indo} have automatically generated a phylogenetic tree for 50 Indo-European languages using a renormalized Levenshtein distance based on a Swadesh list of 200 words that have the same word meanings across languages. We take the Slavic branch from this tree for our similarity analysis. \vspace{0.15cm}

\item  \textbf{Glottochronology-based tree} \hspace{0.1cm} \cite{novotna2007glottochronology} have used a classical glottochronological approach to generate a genetic a tree for the Balto-Slavic languages. Their approach employs a manual calculation of pairwise distances between languages based on the recognition of cognates.    \vspace{0.15cm}

\item  \textbf{Geographic distance-based tree} \hspace{0.1cm} Using the geographic distance matrix we obtained from the ASJP database, we apply the same hierarchical clustering as we applied on our prototype language vectors to obtain a tree based on geographic distance.  \vspace{0.15cm}

\item  \textbf{GLG Confusion-based tree} \hspace{0.1cm} \cite{skirgaard2017some} have applied hierarchical clustering on their player confusion data and their resulting tree shows that the Slavic languages form a pure cluster. We consider this tree to be a good approximation of non-linguist's perception of language variation and similarity.  \vspace{0.15cm}

\item  \textbf{Randomly generated tree} \hspace{0.1cm} To give a reference point for a worst case scenario in our analysis, we generate a tree using the Ward hierarchical clustering on a random confusion matrix. \vspace{0.15cm}
\end{enumerate}

%Reasons why the similarity analysis was conducted on the unweighted trees not matrices  

\noindent
First, and to facilitate the measurement of distance between different trees, we consider Croatian and Serbian as a single node in our tree (i.e., Serbo-Croatian) and keep only the nodes that represent languages shared by all trees. This leaves us with eight languages: Bulgarian, Serbo-Croatian, Slovene, Czech, Polish, Slovak, Russian, and Ukrainian.   Second, we compute the distances between the tree generated from the representation distances of our LID model and each of the aforementioned trees using the unweighted tree distance metric introduced by \cite{rabinovich-etal-2017-found}. The result of this evaluation is presented in Table \ref{tab:tree_dist}. We observe that the most similar tree to our tree is the one based on the player confusion data from the GLG. These findings suggest that the factor that best explains the similarities within the emerging representations from our LID model is the perceptual similarity between Slavic languages, approximated by the confusability obtained from the GLG participants.

\begin{table}[t]
\centering
\begin{tabular}{@{}rc@{}}
\toprule
\textbf{Tree}          & \textbf{Distance} \\ \midrule
Levenshtein distance-based & 0.270               \\
Glottochronology-based    & 0.132               \\
Geographic distance-based  & 0.140               \\
GLG Confusion-based        & \textbf{0.084}                \\ \midrule
\textbf{Random}                & 0.371               \\ \bottomrule
\end{tabular}
\caption{Tree distance evaluation. Lower values correspond to smaller distances. }
\label{tab:tree_dist}
\end{table}

% --------------------------------------------------------------
\section{Discussion}
% --------------------------------------------------------------

Many recent works have shown that deep neural networks are good models of human perception. For example,  \cite{zhang2018unreasonable} and \cite{petersoncapturing}  have shown that emerging representations from neural models trained on visual recognition tasks are predictive of human similarity judgments. For auditory recognition, neural speech recognition models have been shown to capture human-like behavior in cross-lingual phonetic perception \citep{schatz2018neural}. Following the same spirit, our objective is to the investigate the extent to which neural models of spoken language identification capture language similarity. Nevertheless, and because of the complex space in which language variation can be realized, the similarity between two languages is a multidimensional phenomenon that cannot be expressed in a single number \citep{van2008making}. We therefore do not consider a single reference as a ``ground truth'' in our analysis, but consider several reference criteria of distance including genetic, geographic, and perceptual distance.

Our representation similarity analysis shows that emerging representations in our spoken LID model capture language similarity. The representation visualization illustrated in Fig. \ref{fig:viz} demonstrates the generalization ability of our model to project speech segments of non-observed (held-out) languages into subspaces of their respective subgroups. Given that the data in our study constitute contemporary realizations of Slavic speech that do not explicitly encode diachronic sound changes, we first hypothesized that the geographic distances between the linguistic communities would be a good predictor of the distances in the representation space of the LID model. This turns out to indeed be the case as we observe a high positive correlation between geographic distances and cosine distances within prototype language representations. 

On the other hand, we were less optimistic about our LID model capturing the genetic signal between Slavic languages (that is, whether the language representations encode the historical relationships between languages). Earlier works  that have investigated computational approaches to generating genetic language trees have either employed  historical etymological data capturing phonological sound changes \citep{cathcart-wandl-2020-search}, sequences reflecting syntactic patterns in different languages \citep{rabinovich-etal-2017-found, bjerva2019language}, or word lists reflecting lexical similarity \citep{serva2008indo}. Arguably, these sources of language data are more likely to preserve the relationship between languages across the temporal dimension than the contemporary Slavic speech we use in this study. Therefore, our initial intuition was that the resulting tree would reflect variation across the spatial dimension and not the temporal dimension. Nevertheless, the tree generated by our analysis is an adequate approximation of the Slavic genetic tree given the contemporary nature of the data sources. 

Finally, it is striking to observe the similarity between our generated tree and the Slavic subtree constructed from player confusion patterns of GLG participants \citep{skirgaard2017some}. There are two main factors that contribute to this similarity of our analysis to that of \cite{skirgaard2017some}: (1) West-Slavic and East-Slavic branches are clustered together before joining the South-Slavic branch to form a single Slavic cluster, and (2) the deviations from the widely accepted Slavic grouping are present in both analyses at approximately the same node locations since Bulgarian is grouped with East-Slavic and Slovak is grouped with South-Slavic. These findings provide evidence that the perceived language similarity is the factor that can best predict the geometric distance between languages in the emerging representation space from neural models of spoken language identification.  
\section{Conclusion}
% --------------------------------------------------------------
We have presented a convolutional neural model for Slavic language identification in speech signals and analyzed the extent to which its emerging representations reflect language similarity. Our analysis has shown that the distances in the emergent language representations reflect language variation across the temporal (phylogenetic) and spatial (geographic) dimensions. Moreover, by comparing our clustering analysis to the confusion patterns of the Great Language Game participants, we have shown that perceptual confusability is a better predictor of language representation similarities than phylogenetic and geographic distances.

\section*{Acknowledgements}
We would like to thank Nicole Macher for assisting with the research presented in this paper. We extend our gratitude to anonymous reviewers for their insightful suggestions and comments. This research is funded by the Deutsche Forschungsgemeinschaft (DFG, German Research Foundation), Project ID 232722074, SFB 1102.

% The acknowledgements should go immediately before the references.  Do
% not number the acknowledgements section. Do not include this section
% when submitting your paper for review.

%\newpage
\bibliography{coling2020}
\bibliographystyle{plainnat}

\end{document}